# Evaluation of Multimodal Semantic Segmentation using RGB-D Data

Jiesi Hu*[a], Ganning Zhao[a], Suya You[b], C. C. Jay Kuo[a],
[a]University of Southern California, 3551 Trousdale Pkwy, Los Angeles, CA, 90089; [b]Army Research Laboratory, 12025 E Waterfront Dr. Los Angeles, CA, 90094

## ABSTRACT

Our goal is to develop stable, accurate, and robust semantic scene understanding methods for wide-area scene perception and understanding, especially in challenging outdoor environments. To achieve this, we are exploring and evaluating a range of related technology and solutions, including AI-driven multimodal scene perception, fusion, processing, and understanding. This work reports our efforts on the evaluation of a state-of-the-art approach for semantic segmentation with multiple RGB and depth sensing data. We employ four large datasets composed of diverse urban and terrain scenes and design various experimental methods and metrics. In addition, we also develop new strategies of multi-datasets learning to improve the detection and recognition of unseen objects. Extensive experiments, implementations, and results are reported in the paper.

## 1. INTRODUCTION

Semantic scene understanding is crucial for autonomous driving, robot navigation, and intelligent scene perception tasks. Autonomous maneuver in complex environments needs the ability to accurately and robustly distinguish and locate objects in the environments. Semantic segmentation can identify and locate objects at the pixel level, which provides important situation awareness for navigational tasks. Rapid advances in neural networks have led to boost the performance of AI-driven image segmentation models, but the ability to deal with challenging outdoor environments and conditions accurately and robustly is still not sufficient and needed to advance.

Using multi-sensors such as RGB camera, depth camera, and LiDAR sensors can make the perception model more robust under various conditions. For example, the depth sensor can capture 3D distance and geometric information of the environment while the RGB image provides photometric information of the objects and scenes. A hybrid system combing the multiple sensing resources can compensate for the shortcomings of each individual technique by using multiple measurements to produce robust results. The work reported in this paper employs RGB and depth data for semantic segmenting pre-defined objects in challenging urban/terrain environments. The depth data can come directly from a depth camera or from converting 3D point clouds captured with LiDAR sensor or photogrammetry.

When deploying the system, there may be unknown object classes in the pre-defined object training datasets. For example, when driving, we may encounter unexpected obstacles (e.g., trashcan, barrier, rubble, and stones) suddenly appearing on the roads. An autonomous system must be able to recognize and avoid these unseen objects to ensure safety.

Our long-term goal is to develop stable, accurate, and robust semantic scene understanding methods for wide-area scene perception and understanding, especially in challenging outdoor environments. To achieve this, we are exploring and evaluating a range of related technology and solutions, including the AI-driven multimodal scene perception, fusion, processing, and understanding. This work reports our efforts on evaluation methods and results of a state-of-the-art approach for semantic segmentation with multiple RGB and depth sensing data. We consider the case where the autonomous sensing platform is composed of a set of RGB cameras and LiDAR sensors moving unconstrainedly in urban/terrain environments. Under this condition, the data streams of RGB video and LiDAR point clouds are collected in real time and then fed into a downstream system for data preprocessing, calibrating, and converting the 3D point clouds to 2D depth image[30] that results in pair of RGB-D data stream registered in a common coordinate frame. Given the pair of RGB-D data, our segmentation module performs feature detection, data fusion, segmentation, and pixel-level annotation automatically in an end-to-end fashion.

*jiesihu@usc.edu; phone: 1 626 642-6623.

## 2. RELATED WORKS

### 2.1 RGB-D Semantic Segmentation

With the development of depth sensors, people have begun to exploit using both depth and RGB images to obtain better semantic segmentation. Early work such as[7] simply stacked RGB image and depth image together, and then input it into the neural network, but such a structure cannot make full use of the geometric information of object provided by the depth image. Wang[8] proposed Depth-aware CNN to explore depth maps, but this method is only suitable for dense depth maps. FuseNet[9] established two branches to process RGB image and depth image respectively for segmentation in the indoor environment. Park et al.[10] compared the different network structures of feature fusion and proposed feature fusion and refinement blocks. ACNet[20] promoted the performance of the model to a higher level by proposing an attention complementary module to fuse the features of RGB and depth image. These studies have proved that for indoor environment, adding depth map, which encodes rich geometry and structure information of objects and scenes, can help significantly the semantic segmentation model to achieve better performance for indoor scenes.

Different from the indoor environment, the depth map of the outdoor environment is often sparser, and the scenes are more complex. In order to achieve decent segmentation for the outdoor environment, Krešo[11] proposed a scale selection layer to extract the features of RGB image and match them with the reconstructed depth map. Deng et al.[12] proposed a residual fusion block to learn complementary features and extract cross-modal features from multiple inputs. Squeezeseg[13] is a real-time model taking depth map as input and Fuseseg[14] fusing RGB information in it. To obtain better performance, a real-time segmentation network was recently proposed RFNet[6], in which two datasets were used to train the model at the same time. In our work, we leverage more datasets that are collected at scale diverse and realistic variations from different environments, which broadens the application scenarios of the model while ensuring the accuracy and speed of the model.

### 2.2 Unexpected objects detection

For automatic driving, it is very important to identify unexpected objects, because the road conditions are complicated, and detecting unexpected objects on the road may prevent traffic accidents. For this problem, people used geometric information coupled with neural network and visual appearance to complete this task. The geometric point cluster methods proposed in Manduchi[15] and Broggi's[16] works detect and locate obstacle points through the relationship between points. Ramos[17] and Singh[18] obtained semantic prediction of obstacles using CNN. Through multi-source learning, RFNet[6] made the deep learning model be able to detect and recognize many common objects small obstacles in the urban environment. However, the small obstacle samples learned by previous work mostly come from one dataset, which makes it difficult to tackle complex and challenging environments. For possible unseen obstacles encountered in the scene, we create an additional dataset (called *supplemental* dataset) labeled with these obstacles supplementing to the main datasets. Our model employs a multi-dataset learning strategy that can greatly extend the object classes can be recognized and inferred in complex outdoor environments.

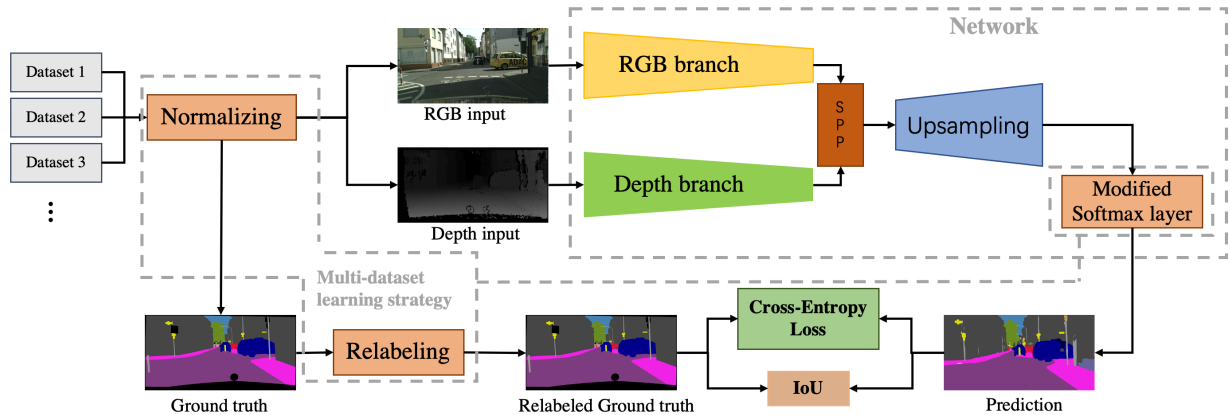

Figure 1. A high-level illustration of the proposed semantic segmentation pipeline

## 3. METHODOLOGY

Figure 1 is a high-level illustration of the proposed semantic segmentation pipeline. First, the inputs of RGB image and depth map are passed through the Normalizing module to resize all the input images to a pre-defined size. Then, the depth map and RGB image are sent to the processing Network, which is composed of RGB branch, Depth branch, SPP, and Upsampling modules. Ground truth will go to Relabeling module. A Multi-dataset learning strategy is applied to Resizing, Relabeling, and Modified Softmax layer. Finally, by comparing the results of relabeled ground truth and segmentation prediction, we obtain the intersection of union (IoU) and value of cross-entropy loss over the end-to-end semantic segmentation network.

### 3.1 Network architecture

The network architecture is built upon the RFNet architecture[6] that consists of three major parts: Decoder, Spatial Pyramid Pooling (SPP), and Encoder.

The Encoder has two branches, which extract features of RGB image and depth map respectively. Both branches use ResNet-18[22] as the backbone because ResNet-18 not only provides good feature extraction ability but also has the decent speed to ensure real-time processing. In order to fully integrate the RGB image and depth map, the output features from each ResNet Blocks are fused together. Each ResNet Blocks contains four convolutional layers in ResNet-18. During the fusing, features from two branches first pass through the Squeeze-and-Excitation block (SE)[27] and then are combined together through element-wise addition. At the end of the decoder, the features of two branches are fused to form fused features inputting into the SPP layer.

It is important to apply the SE block because it can highlight important channels while suppressing unimportant channels in the feature map based on global information. To capture the global information, the feature map that enters the SE block will first do global pooling and then go through a 1×1 convolution layer and sigmoid function to get the weight of each channel. The output of the SE block is obtained by doing the outer product for the weight vector and input feature maps. After adding the SE blocks from both branches element-wise, we can get the fusion feature maps. The SE block can be expressed by the following formula:

$$F = X \otimes \sigma(\phi(X)) \quad (1)$$

where $X$ and $F$ denote the input and output of the SE block respectively. $\phi$ represents global pooling, and $\sigma$ denotes a combination of 1×1 convolution layer and sigmoid function. $\otimes$ stands for the outer product.

In order to increase the visual field of feature for each pixel while ensuring real-time processing, a spatial pyramid pooling (SPP)[25] which outputs the average with multi-scale information is added between decoder and encoder. The input feature of SPP is 32 times smaller than the original image. For example, when 1024×2048 is the original size, the input size of SPP becomes 32×64. After averaging in SPP layer, the heights of the feature maps become 8, 4, and 2 and the width will change in the same proportion. Finally, the sizes of the feature map become the same as the input through bilinear interpolation, and then, all feature maps including the input are concatenated together to form the output of SPP.

The Decoder takes high-level features from SPP module as input. Our decoder is composed of three up-sampling and convolution modules. The up-sampling modules use bilinear interpolation to increase the feature size. The output of the up-sampling module is element-wise added with the features from the skip connection. The value of skip connection is obtained from the output of the corresponding block in the RGB branch through a 1×1 convolutional layer. It can provide detailed information from early layers. After each element-wise adding, we add a convolutional layer to further process the features. The output of the last convolutional layer has the size of ($n$×$h$×$w$), where $n$ is the number of classes. $h$ and $w$ denote the height and width of the feature map respectively and are equal to the size of input images. The Softmax layer outputs a probability distribution per pixel with the form:

$$a_L = \frac{e^{Z_L}}{\sum_i e^{Z_i}} \quad (2)$$

$Z_i$ is the unbounded feature from the last convolutional layer corresponding to class $i$. $a_L$ denotes the probability of class $L$. For each pixel, the class with the highest probability is selected as the prediction. Then we can leverage ground truth and prediction to calculate loss and metric.

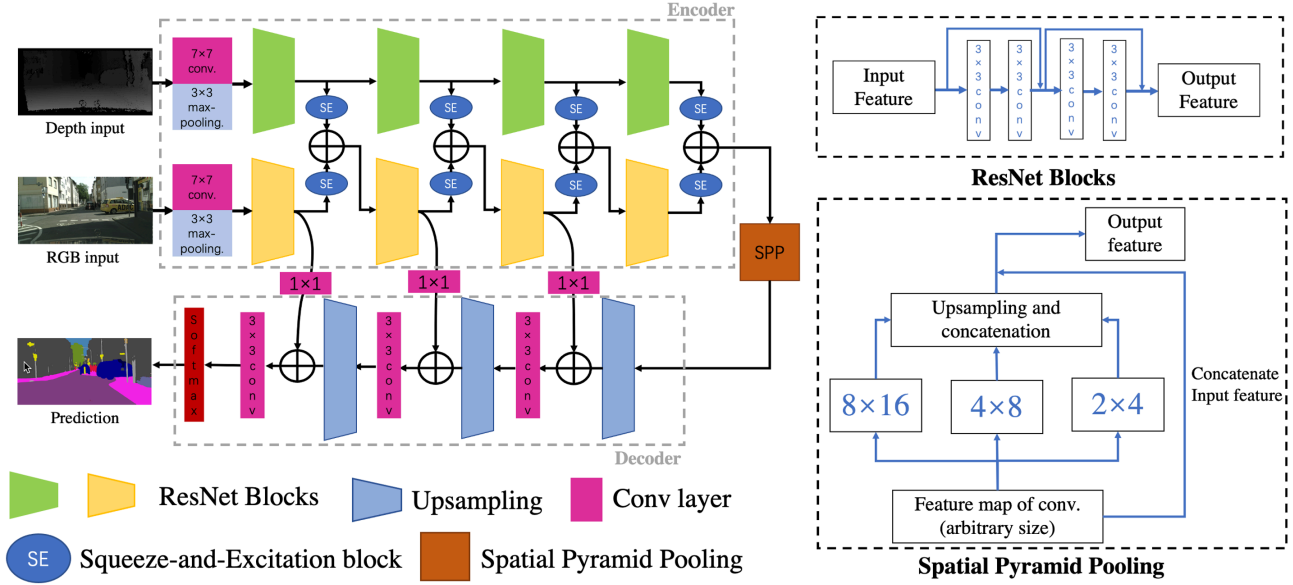

Figure 2. Network architecture and module details

### 3.2 Multi-Dataset learning strategy

In order to cope with the complex conditions in reality and train the neural network model with scale diverse data, we apply the multi-dataset learning strategy. Multi-dataset learning also enables our model to learn more classes including the pre-defined known objects and unseen objects. However, we cannot simply merge the multiple datasets together, since the semantic labels, image sizes and index of different datasets are often incompatible with each other. To solve this issue, we proposed the following methods.

#### 3.2.1 Standard relabeling method

The standard relabeling method has two steps: choose the main dataset among all supplemental datasets and merge the datasets together with defined rules.

For the first step, the choice of the main dataset is determined by the desired class set. The class set of the main dataset will be remain, and for other supplemental datasets, their integrity of the class set cannot be guaranteed. In this paper, we have four datasets: Cityscapes[4], Lost and Found[5], KITTI semantic segmentation[3], and RELLIES-3D[2] datasets. If we want to keep all the object classes in Cityscapes, and the main dataset will be Cityscapes.

In the second step, we will merge the main dataset with other one by one. Let's say D1 is the main dataset, and we want to merge D1 with dataset D2. Assume the class set of D1 is $S^{[1]} = \{s_1^{[1]}, s_2^{[1]}, \ldots, s_i^{[1]}, \ldots\}$, and the class set of D2 is $S^{[2]} = \{s_1^{[2]}, s_2^{[2]}, \ldots, s_j^{[2]}, \ldots\}$,. When merging, consider the elements in $S^{[2]}$ in turn. If $\exists s_i^{[1]}$ in $S^{[1]}$, $s_j^{[2]} \subseteq s_i^{[1]}$, the label corresponding to $s_j^{[2]}$ is changed to that of $s_i^{[1]}$, so these two classes are merged into one class. If $\forall s_i^{[1]}$ in $S^{[1]}$, $s_j^{[2]} \cap s_i^{[1]} = \emptyset$, $s_j^{[2]}$ will be added into $S^{[1]}$ as a new class. In any other cases, $s_j^{[2]}$ conflicts with other classes and will be ignored. This brings together the classes of D1 and D2. We repeat above steps to merge all the datasets.

The loss function for the multi-dataset learning strategy is defined in (3), where $\phi$ is our segmentation model. $L$ denotes cross-entropy loss function. $x_i$ and $y_i$ denote the multi-source training samples and their labels respectively.

$$loss = \frac{1}{l}\sum_{1}^{l} \lambda_i L(\phi(x_i), y_i) \quad (3)$$

$\lambda_i$ is used to determine whether to ignore the training sample, which is defined as follows:

$$\lambda_i = \begin{cases} 1, & Remain\ x_i \\ 0, & Ignore\ x_i \end{cases} \quad (4)$$

For example, when the Cityscapes is the main dataset and we want to merge it with RELLIES-3D, the class "*Tree*" in RELLIES-3D will be labeled as "*Vegetation*" in Cityscapes, because the definition of the Vegetation class in Cityscapes includes "*Tree*". Class "*Puddle*" in RELLIES-3D will become an additional class because Cityscapes has no class related to it. The "*Vehicle*" in RELLIES-3D will be discarded. The reason is that the "*Car*", "*Truck*" and "*Bus*" classes in Cityscapes are all vehicles, so we cannot determine which one the "*Vehicle*" in RELLIES-3D belongs to.

### 3.2.2 Thrifty relabeling method

The scheme of the Thrifty relabeling method is almost the same as the standard relabeling method. The only difference is that when encountering conflict class, it will not discard the conflict classes, but will add the conflict class to the training set as an additional class. In our experiment, the "*Unknown*" class in the Lost and Found dataset set is a conflict class. Using the Thrifty relabeling method, the class Unknown will be treated as an extra class when training the model. However, during testing, we don't want the model to recognize any object as unknown. Thus, the Unknown class will be removed from the class set to ensure reasonable prediction results. Specifically, when testing the model, we set the value of the node corresponding to the conflict classes in the Softmax layer to 0. In this case, we modified the formula of Softmax layer as shown below.

$$a_L = \begin{cases} \frac{e^{Z_L}}{\sum_i e^{Z_i}}, & L \notin S_{conflict} \\ \frac{\eta e^{Z_L}}{\sum_i e^{Z_i}}, & L \in S_{conflict} \end{cases} \quad (5)$$

where $a_L$ denotes value corresponding to class $L$ in the Softmax layer and $S_{conflict}$ is the set of conflict classes. $Z_L$ represents the value of the node corresponding to class $L$ from the last layer. The value of $\eta$ equal to 1 when training the model and becomes 0 when testing. With this strategy, the model gets more data for training and is more likely to extract better features.

### 3.2.3 Resizing methods

The difference in image resolution among datasets brings difficulties to multi-dataset training. In order to cope with this problem, we tested three different resizing methods: *Original Size*, *Same Width*, and *Size Warping* methods. When using the Original Size method, images from all the datasets will not be changed. Since the segmentation model does pixel-wise prediction, the model can still be trained. However, if the resolution gap between the datasets is too large, the big difference in image size may exacerbate the model's performance. For the Same Width method, all the images have the same width while the height is adjusted to keep the original ratio of the images. Through this method, images of different data sets will have similar sizes. When using the Size Warping method, images from all datasets will be warped into the same size using bilinear interpolation. Table 1 shows our setting for the image sizes of different datasets and different resizing methods. We use the image size of the Cityscapes dataset as the standard.

Table 1. The image size of different datasets with different resizing methods. City., Lost., and KITTI. are shorthand for Cityscapes, Lost and Found and KITTI semantic segmentation datasets respectively. All the tables throughout the paper follow the same naming rule.

|  | City. | Lost. | KITTI. | RELLIS-3D |
|---|---|---|---|---|
| *Original Size* | 1024×2048 | 1024×2048 | 347×1241 | 1200×1920 |
| *Same Width* | 1024×2048 | 1024×2048 | 622×2048 | 1280×2048 |
| *Size Warping* | 1024×2048 | 1024×2048 | 1024×2048 | 1024×2048 |

# 4. EXPERIMENTS

## 4.1 Datasets

In this work, four RGB-D semantic segmentation datasets are exploited, including Cityscapes, Lost and Found, KITTI, and RELLIS-3D datasets.

Cityscapes[4] provides data that facilitates the realization of semantic segmentation for the urban environment. Its training, testing, and validation set contain 2975, 500, and 1525 images respectively, whose resolution is 2048 × 1024. The dataset contains 19 different classes, and the depth maps are provided.

The Lost and Found dataset[5] consists of 814 training samples and 1200 validation samples. It has both RGB images and corresponding depth maps. The data is extracted from 112 stereo video sequences, and only fine annotations for road and small obstacles are provided. The resolution of the image is also 2048 × 1024.

KITTI semantic segmentation dataset[3] consists of 200 semantically annotated samples extracted from KITTI video sequences. In our experiment, we split it into 50 validation samples and 150 training samples. It provides the corresponding depth map and fine annotation which is compatible with the Cityscapes dataset. However, its image size is much smaller which is 1241 × 347.

RELLIS-3D[2] provides a large number of terrain scenes and is a supplement to the off-road environment. The annotated dataset includes 3302/983/1267 samples for training, validation, and testing. RGB images, calibrated point cloud, and fine annotations are provided. The depth map can also be generated directly from point clouds. The dataset contains 18 different classes, and the size of the image is 1920 ×1200. Since its validation set only has few pixels of water, object, asphalt, and building objects, these classes are ignored when evaluating on its validation set.

## 4.2 Implementation Details and Metric

We implement the framework in PyTorch and use an NVIDIA Tesla V100 GPU for run-time evaluation. We follow the same training protocol of RFNet[6]. The parameters of layers from ResNet-18 are initialized with ImageNet pre-trained weight. We have achieved a 30.3 FPS inferring rate on our machine. The depth maps of Cityscapes and Lost and Found datasets are obtained by using a semi-global matching (SGM) algorithm[28], while the depth maps of KITTI and RELLIS-3D datasets are generated from LiDAR point clouds. We apply a window with a size of 7 and a stride of 1 to do max pooling on the projection of the point clouds and add zero padding to ensure that the size of depth map is unchanged. When dealing with Cityscapes and Lost & Found datasets, we crop the edge regions and resize images back to the original size, because the edges of their depth map derived from the SGM algorithm are not applicable. For RELLIS-3D and KITTI datasets, we do not crop the images. Examples of inputs are shown in Figure 3.

We evaluate our model's performance on class-level segmentation and compute the IoU (intersection-over-union) and mean IoU (mIoU) scores, which are defined as follows:

$$\text{IoU}_c = \frac{|\mathcal{P}_c \cap \mathcal{G}_c|}{|\mathcal{P}_c \cup \mathcal{G}_c|}, \text{mIoU} = \frac{1}{n}\sum_{c=1}^{n}\text{IoU}_c \quad (6)$$

where $\mathcal{P}_c$ and $\mathcal{G}_c$ denote the predicted and the ground truth set of class-c respectively. $|\cdot|$ denotes the cardinality of the set. n denotes the total number of classes and mIoU is the primary metric in our experiments to compare different models.

## 4.3 Model evaluation with different resizing methods

To fully evaluate the influence of different resizing methods, we compare the models trained with different datasets. Table 2 shows the mIoU of our models in different conditions. The training set and validation set are not overlapped to ensure the accuracy of the evaluation.

It can be seen from Table 2 that for different dataset selections, it is difficult to identify which resizing method is always performing superiorly. Many factors affect the choice of resizing method, including main dataset, training set, and even validation set. Therefore, we cannot conclude which resizing method is the best.

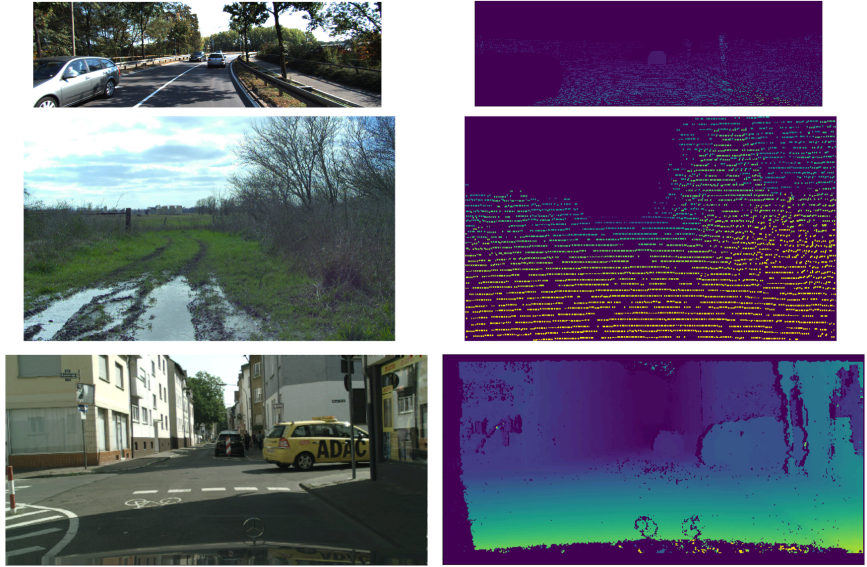

Figure 3. Examples of RGB images and corresponding depth maps from KITTI (top), RELLIS-3D (middle), and Cityscapes (bottom) datasets

Table 2. The mIoU of different datasets with different resizing methods. Four Datasets means using Cityscapes, Lost & Found, RELLIS-3D, and KITTI semantic segmentation datasets together. The dataset in the parentheses is the choice of the main dataset. Table 4 follows the same naming rule.

| Training set | Four Datasets (City.) | | Four Datasets (RELLIS-3D) | | City. + Lost. + KITTI(City.) | |
|---|---|---|---|---|---|---|
| Validation set | Four datasets (City.) | City. + Lost. | Four datasets (RELLIS-3D) | RELLIS-3D | City. + Lost. + KITTI | City. + Lost. |
| Original Size | 63.79% | 72.10% | 60.88% | **49.11%** | 71.03% | 71.44% |
| Same Width | **67.14%** | **72.29%** | 62.56% | 48.05% | 71.14% | 71.54% |
| Warping | 65.42% | 71.71% | **63.75%** | 48.61% | **71.65%** | **71.69%** |

### 4.4 Evaluation of model structure and multi-dataset learning strategy

We perform an ablation study to evaluate model structure and multi-dataset learning strategy. In Table 3, all methods are modified based on RFNet architecture. The Single RGB method only establishes the RGB branch to perform semantic segmentation using RGB input. RGB-D-Stack combines RGB image and depth map to form a 4-channel input, and only uses one branch for segmentation. Through this method, we try to find out whether two branches should be built to extract the features of the depth map and RGB image respectively. RGB-D-Fusion (concatenation) has two branches to extract the features of the RGB image and depth map, but it applies concatenation to fuse the feature map instead of element-wise adding. RGB-RGB-Fusion (element-wise add) replaces the depth input with RGB input to figure out whether the information of the depth map can benefit semantic segmentation. RFNet (thrifty) and RFNet (standard) both apply the model described in Section 3. The difference is that RFNet (thrifty) adopts the thrifty relabeling method, while RFNet (standard) uses the standard relabeling method.

In general, there are four cases in this table and different data is leveraged in each case. For different datasets, we chose the best performing resizing methods to do resizing. First, let's consider the case of using only RELLIS-3D terrain dataset. Since the training set only has one dataset, the mIoU of RFNet (thrifty) and RFNet (standard) are the same. RGB-RGB-Fusion (element-wise add) has a similar performance with RFNet (standard), which shows that the depth map is not very helpful. RGB-D-Fusion (concatenation) gets the best performance in this case, indicating that the network architecture applied to the city is not the best for the terrain scene. When the training set is City + Lost., the accuracy of RFNet (thrifty) is 72.81% exceeding all other methods. It is worth noticing that compared with RFNet (standard), RFNet (thrifty) only changes the way of combining multiple datasets and does not change the network architecture. For the model using City + Lost + KITTI datasets, the performance of other methods is far lower than RFNet and the result of RFNet (thrifty) is still the best. In the case of Four Datasets (Cityscapes), the validation set includes both urban and terrain environments. RFNet (standard) gets the best performance, followed by RFNet (thrifty). Their results are both obviously higher than other methods.

In summary, the results show that the performance of RFNet based model is much better than other methods for urban and blended environments. RFNet (thrifty) has the best performance in two of the four cases and its performance in other cases is very close to the best one.

Table 3. mIoU of models with different datasets and designed method choices

| Training data | RELLIS-3D | City. + Lost. |
|---|---|---|
| Validation data | RELLIS-3D | City. + Lost. |
| Single RGB | 49.98% | 69.20% |
| RGB-D-Stack | 49.61% | 65.20% |
| RGB-D-Fusion (concatenation) | **54.35%** | 68.67% |
| RGB-RGB-Fusion (element-wise add) | 54.06% | 69.37% |
| RFNet (standard) | 54.17% | 72.22% |
| **Ours-RFNet (thrifty)** | 54.17% | **72.98%** |
| Training data | City. + Lost. + KITTI. | Four Datasets (Cityscapes) |
| Validation data | City. + Lost. + KITTI. | Four Datasets (Cityscapes) |
| Single RGB | 68.00% | 66.32% |
| RGB-D-Stack | 66.39% | 54.36% |
| RGB-D-Fusion (concatenation) | 68.31% | 54.74% |
| RGB-RGB-Fusion (element-wise add) | 69.82% | 65.29% |
| RFNet (standard ) | 71.65% | **69.94%** |
| **Ours-RFNet (thrifty)** | **72.12%** | 68.03% |

## 4.5 Performance of models on commonly used validation sets

In this experiment, the Cityscapes dataset is chosen to be the main dataset of all models. In Table 4, we report mIoU of RFNet with different numbers of training datasets and relabeling methods. Their relabeling methods are reflected in their name and their corresponding training sets are shown in the table. Another famous real-time network is listed: SwiftNet[23]. For a fair comparison, SwiftNet is retrained in our machine with the same multi-dataset learning strategy.

Table 4. Comparison of different models with same validation sets

| Model | Training set | | | | Validation set | |
|---|---|---|---|---|---|---|
| | City. | Lost. | KITTI. | RELLIS-3D | City. + Lost. | City. |
| SwiftNet-2datasets-standard | √ | √ | | | 70.35% | 72.31% |
| RFNet-2datasets-standard | √ | √ | | | 72.22% | 74.01% |
| RFNet-2datasets-thrifty | √ | √ | | | **72.98%** | **74.76%** |
| RFNet-3datasets-standard | √ | √ | √ | | 71.69% | 73.43% |
| RFNet-3datasets-thrifty | √ | √ | √ | | 72.48% | 74.37% |
| RFNet-4datasets-standard | √ | √ | √ | √ | 72.29% | 73.90% |
| RFNet-4datasets-thrifty | √ | √ | √ | √ | 71.93% | 73.56% |

In Table 4, the performance of RFNet-2datasets-thrifty is the best among all methods for both City. + Lost. and City. validation set. In addition, when the number of training sets increases, the training data and the validation set are no longer strictly corresponding to each other, because different resolutions and scenarios are added into the training set. However, the mIoU is only reduced slightly after more training sets are added, which demonstrates the powerful learning capabilities and robustness of our model. Table 5 shows the per-class IoU of RFNet-2datasets-standard and RFNet-2datasets-thrifty. The only difference between these two models is the relabeling method. In Table 5, RFNet-2datasets-thrifty has obvious advantages in class Sidewalk, Traffic Light, Truck, Bus, and Train. It proves that the conflict class abandoned by the standard relabeling method still provides some useful information, and the thrifty relabeling method can make good use of them.

Table 5. Per-class IoU(%) results of two networks on the blended validation set of City. and Lost.

|  | road | sidewalk | building | wall | fence | pole | traffic light | traffic sign | vegetation | terrain |
|---|---|---|---|---|---|---|---|---|---|---|
| RFNet-2datasets-standard | 95.98 | 60.56 | 90.77 | 50.24 | 59.91 | 60.01 | 62.58 | 72.76 | 91.06 | 57.28 |
| RFNet-2datasets-thrifty | 96.90 | 66.66 | 91.18 | 49.36 | 58.93 | 60.15 | 63.90 | 72.05 | 90.96 | 58.00 |
|  | sky | person | rider | car | truck | bus | train | motorcycle | bicycle | small obstacles | mIoU |
| RFNet-2datasets-standard | 92.54 | 76.15 | 57.88 | 93.32 | 73.78 | 82.27 | 73.17 | 53.98 | 72.22 | 67.90 | 72.22 |
| RFNet-2datasets-thrifty | 92.80 | 76.14 | 59.48 | 93.62 | 76.18 | 83.98 | 76.11 | 54.34 | 72.73 | 66.04 | 72.98 |

### 4.6 Performance of models on validation sets of multiple datasets

To explore the benefit of adding multiple datasets, we report the performance of three models mentioned in Table 6 on validation sets of Cityscapes, Lost and Found, KITTI semantic segmentation, and RELLIS-3D. Models are named in the same way as Table 4.

In Table 6, RFNet-2datasets-standard only performs well on City. and Lost., and worse on other datasets. RFNet-3datasets-standard performs better on KITTI., but it still performs poorly on RELLIS-3D. The RFNet-4datasets-standard has a good performance on all datasets. Figure 4 shows the corresponding visual result which matches the data in Table 6.

Table 6. mIoU(%) on the validation set of Cityscapes + Lost and Found, KITTI semantic segmentation, and RELLIS-3D dataset

|  | Validation set | | |
|---|---|---|---|
|  | City. + Lost. | KITTI. | RELLIS-3D |
| RFNet-2datasets-standard | 72.22% | 23.35% | 14.54% |
| RFNet-3datasets-standard | 71.69% | 52.53% | 11.65% |
| RFNet-4datasets-standard | 72.29% | 60.43% | 62.21% |

Table 7. Changes of class set after training with more datasets

|  | **Original classes** | **New classes** | | |
|---|---|---|---|---|
|  | City. (main dataset) | Lost. | KITTI. | RELLIS-3D |
| Class set | road, sidewalk, building, wall, fence, pole, traffic light, traffic sign, vegetation, terrain, sky, person, rider, car, truck, bus, train, motorcycle, bicycle | small obstacle | (None) | water, object, log, barrier, puddle, rubble |

Table 7 reports the changes in the class set that the model can identify. The Cityscapes dataset provides 19 common classes. After adding the Lost and Found dataset, the model can additionally identify Small Obstacle. The third row of Figure 4 shows that the model can recognize the cargo locating in the middle of the road, which benefits from the Lost and Found dataset. KITTI semantic segmentation dataset does not provide any new class because it has the same class set as the Cityscapes dataset. After joining RELLIS-3D, the number of recognizable classes has increased by six, namely Water, Object, Log, Barrier, Puddle, Rubble. This helps the model to better identify hazards in the terrain. The last two columns in Figure 5 show that after adding RELLIS-3D data, the RFNet-4datasets-standard model can accurately mark hazards such as rubble on purple and barrier on blue.

In general, a model with fewer datasets has poorer adaptability to images of different environments and resolutions. On the contrary, RFNet-4datasets-standard trained with multiple datasets is capable of recognizing more objects and adapting both urban and terrain environments. The result reflects the power of the multi-dataset learning strategy and the potential of RFNet. In addition, although the depth maps from LiDAR and the SGM algorithm have different characteristics, our model can adapt to both of them simultaneously.

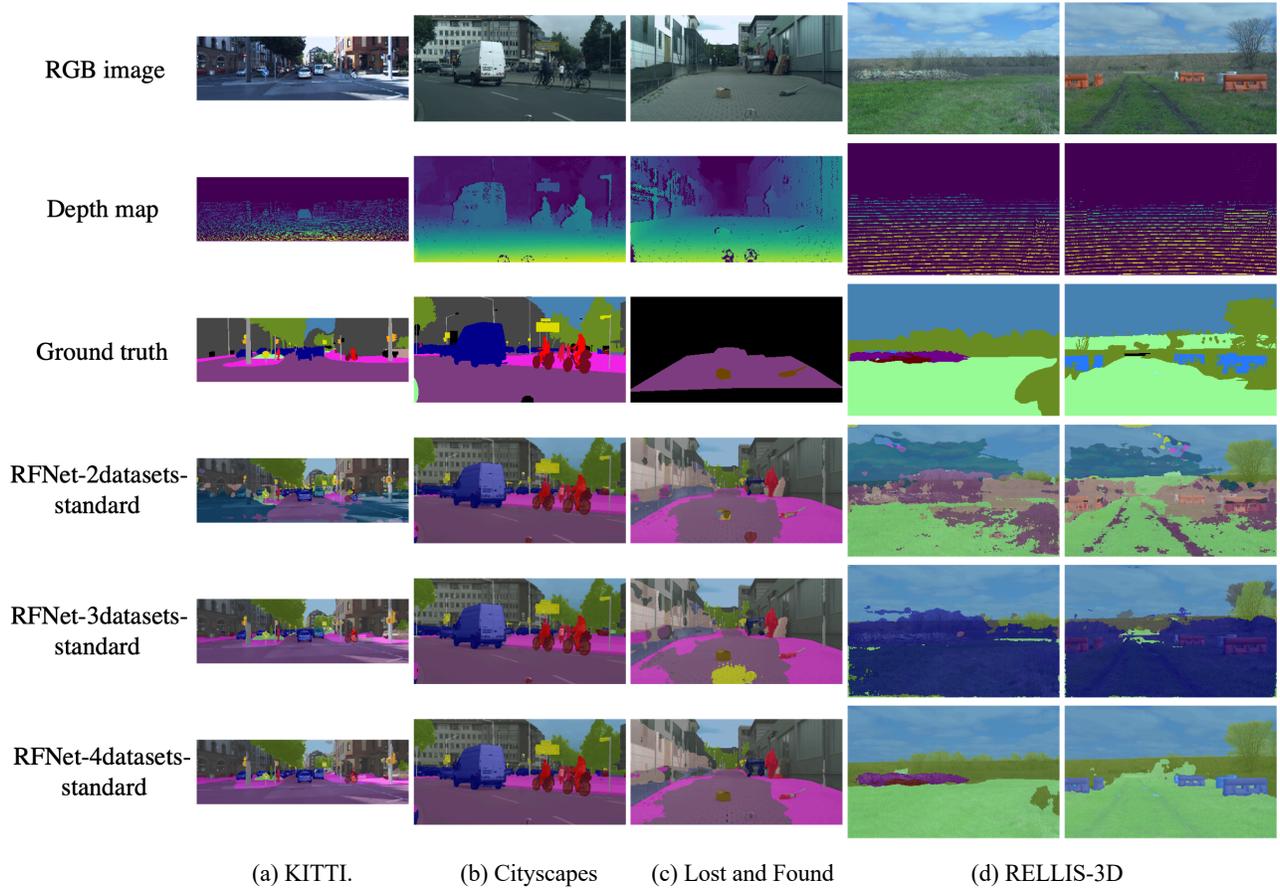

Figure 4. Semantic segmentation results of RFNet-2datasets-standard, RFNet-3datasets-standard, and RFNet-4datasets-standard on various datasets with corresponding RGB images, depth maps, and ground truths

## 5. CONCLUSION

We extensively evaluated state-of-the-art methods for semantic segmentation with multiple RGB and depth sensing data. This paper reported the evaluation methods and results. We employed four different large datasets composed of diverse urban and terrain scenes. We designed different experimental methods and metrics including resizing and relabeling methods. We also proposed new strategies of multi-dataset learning to improve the detection and recognition of unseen objects. We designed and implemented the entire pipeline and network models based on the RFNet architecture to perform semantic segmentation over the designed datasets and scenarios. Our current methods and design are based on the conventional deep neural network (DNN) architecture, which has exposed several inherent problems of DNNs. Our future work will explore a novel non-DNN, non-backpropagation approach[29] for multimodal semantic segmentation.

## ACKNOWLEDGMENTS

This work is supported by US Army Artificial Intelligence Innovation Institute (A2I2). Computation for the work was supported by the University of Southern California's Center for High Performance Computing (carc.usc.edu).